# Similarity-based prediction of Ejection Fraction in Heart Failure Patients

Jamie Wallis[*,1], Andres Azqueta-Gavaldon[*,1], Thanusha Ananthakumar[1]
, Robert Dürichen[+,1], Luca Albergante[1]

[1] Sensyne Health Plc, Oxford, UK
[*] These authors contributed equally to this work
[+] Correspondence to robert.durichen@sensynehealth.com

Abstract:

*Biomedical research is increasingly employing real world evidence (RWE) to foster discoveries of novel clinical phenotypes and to better characterize long term effect of medical treatments. However, due to limitations inherent in the collection process, RWE often lacks key features of patients, particularly when these features cannot be directly encoded using data standards such as ICD-10. Here we propose a novel data-driven statistical machine learning approach, named Feature Imputation via Local Likelihood (FILL), designed to infer missing features by exploiting feature similarity between patients. We test our method using a particularly challenging problem: differentiating heart failure patients with reduced versus preserved ejection fraction (HFrEF and HFpEF respectively). The complexity of the task stems from three aspects: the two share many common characteristics and treatments, only part of the relevant diagnoses may have been recorded, and the information on ejection fraction is often missing from RWE datasets. Despite these difficulties, our method is shown to be capable of inferring heart failure patients with HFpEF with a precision above 80% when considering multiple scenarios across two RWE datasets containing 11,950 and 10,051 heart failure patients. This is an improvement when compared to classical approaches such as logistic regression and random forest which were only able to achieve a precision < 73%. Finally, this approach allows us to analyse which features are commonly associated with HFpEF patients. For example, we found that specific diagnostic codes for atrial fibrillation and personal history of long-term use of anticoagulants are often key in identifying HFpEF patients.*

1. Introduction:

Heart failure (HF) is medical condition characterized by the inability of the heart to pump enough blood throughout the body. Estimates suggests that one in five men and women will develop HF in their lifetime (Lloyd-Jones *et al.*, 2002), resulting in over 8 million people living with HF by the year 2030 in the United States (US) (Heidenreich *et al.*, 2013). While a cure is not currently available, therapies developed to manage HF have successfully decreased mortality and improved the quality of life. This results in projected direct medical costs doubling in the next 20 years to $53 billion in the US alone (Heidenreich *et al.*, 2013). Similar trends are projected worldwide (Heidenreich *et al.*, 2013).

The current clinical practice classifies HF patients into groups based on ejection fraction (EF) (i.e. the volumetric percentage of blood ejected from the left ventricle of the heart during each contraction) with lower values of EF being associated with a more severe disease: heart failure with *reduced* EF (HFrEF, with EF <= 30%), heart failure with *mildly reduced* EF (HFmrEF, with 30% < EF <= 50%), and heart failure with *preserved* EF (HFpEF, with EF > 50%) (Simmonds *et al.*, 2020). To this date, limited research is available on HFpEF patients compared to the more severe HFrEF and HFmrEF patients (Borlaug, 2020) with analyses often focused solely on descriptive statistics (Lee *et al.*, 2009). This can to some extent be explained by the fact that HFpEF does not have its own ICD10 code. Some ICD 10 codes are more likely to capture HFpEF such as I50.3: Diastolic (congestive) heart failure. Although alternative classification formats might include specific diagnoses of HFpEF (such as SNOMED CT) these are not always accessible. Key challenges in the treatment of HFpEF patients include their identification, and the usage of an appropriate treatment regimen since, given the milder presentation of the disease, the side effect associated with medications used to treat the condition may overshadow the benefits.

Clinical Real Word Evidence (RWE), particularly electronic patient records (EPR), provides a great opportunity to better characterise different groups of HF patients and to explore long-term effects of different medications across patients with a range of comorbidities (Blecker *et al.*, 2016; Tison *et al.*, 2018). Unfortunately, due to the way in which data are collected and recoded, information on the patients can be missing, e.g., due to data being stored in databases not fully linked to EPR or data being collected by different non-interconnected clinics (Koye *et al.*, 2020; Bloom *et al.*, 2021). While several imputation schemes are available to researchers (Wells *et al.*, 2013; Beaulieu-Jones and Moore, 2017; Li *et al.*, 2021), the complex nature of clinical data, class imbalance and missingness biases often limit the applicability of standard methods.

In this article we present *Feature Imputation via Local Likelihood* (FILL), a novel methodology designed to impute the value of a specific feature by exploiting available data while implicitly correcting for the presence of different unknown groups of patients which may be associated with different mechanisms leading to the value to be imputed. Two easily interpretable parameters, described later, can be used to optimise the performance of FILL. Note that FILL is not designed to impute values for *all patients*, but to *identify patients with a high likelihood of a correct imputation and perform imputation only on those patients*.

In this article we show how FILL can be used to identify patients with a high-likelihood of having HFpEF in secondary care RWE datasets provided by two NHS Trusts. The value of FILL is further supported by comparing its performance to commonly used ML approaches.

Identifying features associated with HFpEF patients is important not only from an epidemiological perspective, but also to uncover potential prognostic markers and to discover potential novel physio-pathological pathways associated with the disease which may, for example, be used to propose novel drug targets. Previous work on building machine learning algorithms designed to predict if a patients can be classified as HFpEF (Austin *et al.*, 2013; Ho *et al.*, 2016; Uijl *et al.*, 2020) indicates that, despite using a comprehensive set of clinically relevant features, classical algorithms have limited predictive power (See Table 1 and 2). For example, using different alternative methods such as logistic regression, regression trees, support vectors machines and boosted regression trees Austin *et al*. (2013) did not manage to get more than a concordance statistic (c-statistic) of 0.78 (given by the logistic regression model). The c-statistic measures the probability that given 2 individuals (e.g. one with preserved ejection fraction and one with reduced ejection fraction) the model will yield a higher chance for the patient with preserved ejection fraction. It ranges from 0.5 (random concordance) to 1 (perfect concordance) and for binary cases as in ours, the c-statistic is equal to the area under the receiver operating characteristic (ROC) curve (Pencina and D'Agostino, 2015). A similar c-statistic was achieved by Ho *et al*. (2016) using a multivariate logistic regression: 0.80. Furthermore, including a third categorical variable into the predictive variable, mid-range ejection fraction (HFmrEF) and using a multivariable multinomial analysis, Uijl *et al*. (2020) achieved a c-statistic of 0.76 (one versus the rest). This suggests that HFpEF patients are comprised of different subgroups (*phenotypes*), with potentially different driving factors or emphasis of those factors, hence limiting the power of algorithms designed to give similar weight to features across all the patient population. Indeed, several other studies have proposed that HFpEF consists of several different overlapping syndromes which inherently makes it hard to identify (Kao *et al.*, 2015; Uijl *et al.*, 2021).

Table 1: Types of data used to classify HFpEF patients across the literature. Demographics indicates that age and sex of the patient has been used. Examples of Vital signs are beats per minute while examples of laboratory results are systolic and diastolic blood pressure.

| Reference | Data types used | | | | | | Patients Numbers | Performance |
| --- | --- | --- | --- | --- | --- | --- | --- | --- |
| | Demographics | Vital signs | Laboratory results | Medications | Diagnosis | Procedures | | |
| (Uijl *et al*., 2020) | Yes | Yes | Yes | Yes | Yes | No | 10,627 | c-statistic: 0.76 |
| (Ho *et al.*, 2016) | Yes | Yes | Yes | No | No | No | 28,820 | c-statistic: 0.80 |
| (Austin *et al.*, 2013) | Yes | Yes | Yes | No | Yes | No | 4,515 | c-statistic: 0.772 to 0.774 |

FILL complements such approaches in situations when the key aim of an analysis is to *extend* a cohort of HFpEF patients by including other patients with unknown EF status, for which however, there is strong statistical evidence of feature similarity with available HFpEF patients. Furthermore, the working of FILL allows easy explainability for the predictions, hence fostering reproducible results and clinical validation.

From a high-level point of view, when trying to infer a value for an *imputing feature* in a *test patient*, FILL identifies a neighbourhood composed of patients with a known value for the imputing feature which are similar to the test patient across other features. Statistical testing is then used to identify if a specific value for the imputing feature is overrepresented in the selected neighbourhood. If an overrepresented value is present, that value would be assigned to the test patient (see Figure 1).

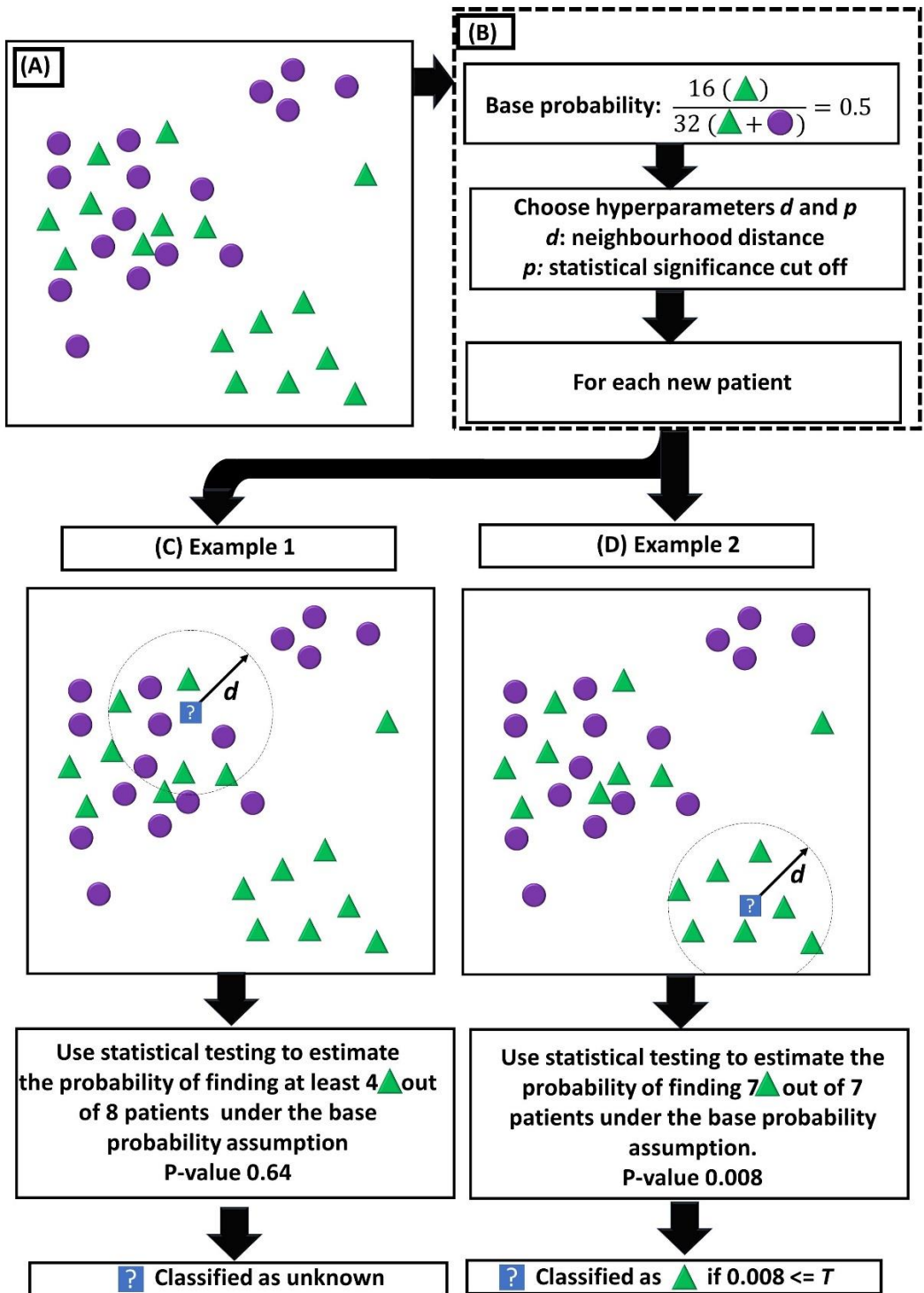

*Figure 1: Flow diagram of the FILL algorithm. Starting from a set of patients with a known value for the feature to be imputed (Panel A) a baseline probability is calculated and hyperparameter optimization can be performed to identify the best neighbourhood distance (d) and statistical significance cut off (p) that maximizes the precision of the algorithm (Panel B). When a patient with an unknown value for the feature to be imputed is considered, they are compared to other patients with similar features (Panels C and D) in order to identify if they are in a neighbourhood with an overrepresentation of a specific value. Panel C shows an example of equal probability of the classes while Panel D shows the extreme case when there is only a single class in a neighbourhood.*

## 2. Materials and Methods:

### 2.1. Data used

This study uses EPR datasets provided by two NHS trusts: Oxford University Hospitals (OUH) and Chelsea and Westminster Hospital (Chelwest). These datasets contain anonymized adult patients with up to 10 years of secondary care data available for each patient. The data comprise of hospital admissions, ICD-10 diagnosis codes, OPCS-4 procedure codes, names of medication prescribed and administered, laboratory measurements, and patient demographics.

Patients were classified as HF patients if they had been assigned any of the following ICD-10 codes: I50* ("heart failure"), I11.0 ("hypertensive heart disease with congestive heart failure"), I13.0 ("hypertensive heart and renal disease with congestive heart failure"), or I13.2 ("hypertensive heart and renal disease with both congestive heart failure and renal failure") at any time in their history. Whilst, the ICD-10 code I50.1 (heart failure with left ventricular failure) should exclusively identify HFpEF patients, several patients with this diagnosis code were recorded as HFpEF in both trusts. This may be due inaccurate recording of either ICD-10 code or EF status. Therefore, patients assigned this diagnosis code were included in our analysis despite the ground truth that, given perfect recording and data quality, no HFpEF patient should have an assignment of the ICD-10 code I50.1. This resulted in 11,950 patients at OUH and 10,051 patients at Chelwest with 2,418 and 1,771 patients, respectively, with known EF measurements. Patients may have multiple EF measurements in their EPR, therefore EF measurement were aggregated to the lowest measurement recorded, i.e., if a patient ever had a HFrEF measurement they were assigned an EF status of HFrEF. This is in line with clinical practice whereby a patient who has ever been recorded as HFrEF should always be treated as a HFrEF patient, regardless of recovered EF status (Wilcox *et al.*, 2020). Of the 2,418 patients in OUH with recorded EF status, 875 had an EF > 50% (which we will refer to as HFpEF) and 1,543 were HFrEF, while for Chelwest there were 1,079 HFpEF patients and 692 HFrEF patients. A small number of patients (39 in OUH and 241 in Chelwest) with records of HFmrEF were considered with unknown EF due to 1) concerns on their representativeness and 2) the absence of specific treatment guidelines for these patients in the NICE guidelines.

Diagnosis codes could either appear in the data as a primary diagnosis (indicating the primary reason for the hospital admission) or secondary diagnoses (indicating further comorbidities relevant to the hospital admission). Similarly, procedure codes can be either primary (indicating the main description of the procedure) or secondary (indicating further aspects of the procedures such as laterality or techniques). Primary and secondary diagnoses, procedures, medication names and laboratory measurements were used in an aggregated fashion and binarised for their presence/absence in the entire patient history while age was calculated at first record of HF diagnosis.

To improve patient anonymity and avoid effect associated with few very uncommon features, we ignored all diagnoses, procedures, medications, and laboratory measurements that were present in less than 1% or more than 99% of patients with known EF status. For diagnoses, this resulted in 385 and 462 different diagnoses initially included in the analysis from OUH and Chelwest, respectively. Procedures were explored by either exclusively using primary procedures or primary and secondary procedures combined. After filtering, 84 and 118 different OPCS-4 codes remained for OUH and Chelwest, respectively when primary procedures were used. When Primary and secondary procedures were combined there were 196 and 219 different OPCS-4 codes, respectively, after filtering. Two forms of medication names were used: the raw medication name entered in the EPR, and a medication name that had clinically-guided standardisation applied. This standardisation was included to reduce some of the noise introduced by free-text entry into the EPR. Filtering the outliers, as above, resulted in 277 and 479 different medication names (240 and 420 after standardisation, respectively) for OUH and Chelwest, respectively. For laboratory measurements 48 and 270 different measurement types remained for OUH and Chelwest, respectively.

Different combinations of the available data were used to assess an optimal feature set (Figure 2).

### 2.2. Description of algorithm

FILL is a distance-based approach for classification problems that exploit the local neighbourhood of a new data point to determine if the new point can be classified with a high likelihood as a given class, in our case, HFpEF due to that class being predominant in its neighbourhood. The method itself has two hyperparameters: the size of the neighbourhood considered, $S$, and the level of statistical significance, $T$ (Figure 1). The results will also be influenced by the choice of the distance measure used, $D$.

Initially, the base probability of encountering a HFpEF patient is calculated by finding the proportion of all known EF statuses that are HFpEF. Second, within the local neighbourhood of a new patient, the number of HFpEF patients is found. Statistical testing is then used to determine if there is a higher-than-expected number of HFpEF patients within the neighbourhood. The resulting p-value is therefore associated with the probability of observing at least as many HFpEF patients, given the number of known patients within the local neighbourhood and the overall base probability. Lastly, a new patient is labelled as HFpEF if the p-value is smaller than the hyperparameter $T$, else the patient is labelled as unclassified. For this analysis, a one-tailed binomial test has been used to compute the p-value, but alternatives tests can be used in different scenarios (e.g., if more than two classes are present).

### 2.2.1. Description of distance metrics used

In this paper, three distance measures have been tested for calculating the distance between two data points: Jaccard, Manhattan, and Gower. The Jaccard distance (Jaccard, 1912) has been introduced to deal with asymmetric binary data. The Manhattan distance is equivalent to the Hamming distance for binary data. Both Jaccard and Manhattan distances were calculated using the *dist* function as part of the *stats* package in R. The Gower was used to deal with mixtures of binary and continuous data such as age (Gower, 1971). The Gower distance was calculated using the *gower_dist* function from the *gower* package in R. The Gower distance can be seen as a combination of the Jaccard (for binary data) and Manhattan (for continuous data) distance measures. For further information on the distance measures used, see supplementary material.

### 2.2.2. Description of parameter optimisation

In order to assess the model performance, a leave-one-out analysis was performed, whereby, we aimed to predict the EF status of a patient with known EF.

A grid search method was applied to find the optimal values of the two hyperparameter S and T. Two such definitions of optimal were used. The first being the pair of hyperparameters that maximised the model's precision subject to the discovery of at least 10 true positives. The second definition of optimal used was the model which maximises the number of true positives subject to a precision above a specified threshold, which here we set to be 0.85.

### 2.2.3. Model Explainability

To better understand why patients are classified as HFpEF (hence providing model explainability) we first identified patients with known EF status within the neighbourhood of each newly HFpEF classified patient. Then, the distribution of features for these neighbours has been compared to the patients with known EF status outside of the neighbourhood. Binary features were compared using the *fisher.test* function as part of the *stats* package in R which performs Fishers Exact test. Age, being the only continuous feature used, was compared using the *t.test* function as part of the *stats* package in R which performs a two-tailed T-test. P-values were adjusted for multiple comparisons by the *False Discovery Rate* (FDR) using the *p.adjust* function as part of the *stats* package in R (Jafari and Ansari-Pour, 2019). The size of the difference (for a feature) between the neighbouring patients and the non-neighbouring patients is then computed as the odds ratio for binary data and difference in mean for continuous data.

### 2.3. Classical machine learning methods

As baseline models to compare the precision obtained by FILL, we ran two classical machine learning methods: multivariate logistic regression and random forests. Multivariate logistic regression models were analysed using the default (0.5) and the optimal probability cut-off (Chan *et al.*, 2003). Using the optimal cut-off probability has been shown to improve the accuracy when working with imbalanced data (Chan *et al.*, 2003). The multivariate logistic regression was performed using the *glm* function as part of the *stats* package in R while the optimal probability cut-off was determined using the *optimalCutoff* function as part of the *InformationValue* package in R which computes the optimal value that minimises the misclassification error. Random forests were performed using 200 trees and bagging after parameter tuning using the *randomForest* function as part of the *randomForest* package in R.

### 2.4. UMAP projection

The Uniform Manifold Approximation and Projection (UMAP) (McInnes, Healy and Melville, 2020) was created using the *umap* function as part of the umap package in R. Age was normalized to take values between 0 and 1, the size of local neighbours and the dimension of the space to embed into were set to their default 15 and 2 respectively while the metric used was Euclidian (different distance metrics barely changed the outcome).

3. Results:

As a first step to test FILL, we explored how different combinations of input data and distance metrics affect the performance of the algorithm using a leave-one-out approach (Figure 2) in the two Sensyne Health partner trusts (see Methods section). For each combination $i$, we computed the values of $S_i$ and $T_i$ that maximise the precision when predicting HFpEF using all the patients with a known EF status. Using the obtained $S_i$ and $T_i$ we computed the precision of the algorithm, the percentage of true positives, the percentage of false positives, and the number of patients with unknown EF status that would be classified as HFpEF by the algorithm.

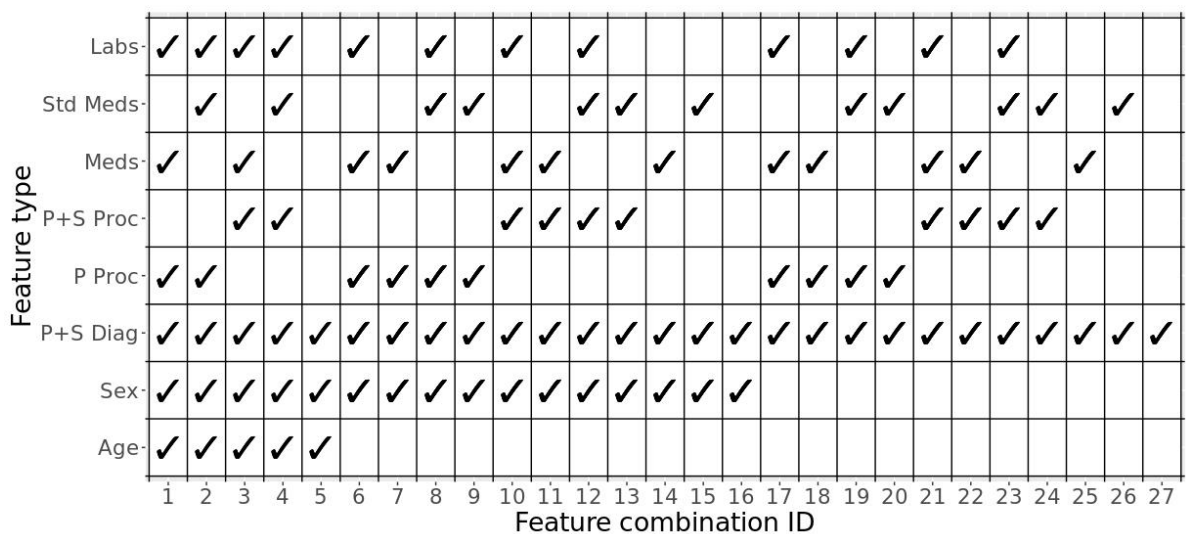

*Figure 2: Feature combinations used in analysis. P+S Diag - Primary and Secondary Diagnoses combined; P Proc – Primary procedures; P+S Proc – Primary and Secondary Procedures combined; Std Meds – Medication names with clinically-guided standardisation. Each combination was tested in the two trusts.*

FILL was able to achieve a precision higher than 0.8 (ranging from 0.803 - 1) in all the scenarios considered (Figure 3A). While classical machine learning methods, such as logistic regression and random forests were only able to achieve precisions below 0.73 (Table 2). Random forests was able to perform marginally better than logistic regression, likely due to non-linearities in the data.

Table 2: Accuracy and precision of logistic regression and random forest in predicting HFpEF. Logistic regression results are presented using the default and optimal cut-off probability. Results using the optimal cut-off probability are displayed in brackets.

| Model and Data | Accuracy | Precision |
| --- | --- | --- |
| | | |
| *Logistic Regression (in brackets results using the optimal cut-off probability)* | | |
| OUH Diagnoses | 0.62 (0.63) | 0.48 (0.49) |
| OUH Diagnoses + Sex | 0.61 (0.62) | 0.45 (0.47) |
| Chelwest Diagnoses | 0.57 (0.57) | 0.68 (0.68) |
| Chelwest Diagnoses + Sex | 0.56 (0.56) | 0.71 (0.71) |
| | | |
| *Random Forest* | | |
| OUH Diagnoses | 0.71 | 0.67 |
| OUH Diagnoses + Sex | 0.70 | 0.67 |
| Chelwest Diagnoses | 0.69 | 0.71 |
| Chelwest Diagnosis + Sex | 0.71 | 0.72 |

As expected, both the trust of origin and the distance measure used influence the precision and the newly HFpEF classified patients (Figure 3). Figure 3-A shows that one of the Trusts (Chelwest) an extremely high precision (~1) is achieved regardless of the distance measure used. The prediction model performed more modestly for OUH data but still managed to achieve a very high precision on average (>0.9). When considering the different distance measures, we observe that Manhattan is overall resulting in the best accuracy, followed by Jaccard and Gower.

While precision is very helpful to evaluate the performance of the algorithm, in most data analysis scenarios the key goal is to increase the number of patients with available measurements. For this reason, we place a key emphasis at the proportion of extra patients (percentage with respect to number of patients with available EF measurements) with unknown EF status being classified as HFpEF (Panel 3-B). For simplicity, we will call this metric, the "*proportion*". In this case, the Gower distance was consistently able to extend the cohort of patients with available EF (proportion) by over 1% regardless of the trust of origin and outperforming the Manhattan distance also despite the trusts (Figure 3-B). This is not unsurprising and suggests that higher accuracy is achieved by assigning HFpEF to a smaller number of patients for which there is more data support. The interquartile range (IQR) associated with the boxplots also indicates that the proportion changes significantly depending on the data used (diagnosis, medication, procedures or LIMS).

All in all, our analysis indicates that the Gower distance seems to be preferable as it results in consistently large proportions (Figure 3-B) while maintain a high precision (Figure 3-A)

To further assess the performance of the different combination used (see Figure 2), Figure 3-C shows the proportion of new patients identified across distances (shape) and trust (colour) for each of the 27 combinations. Of note is combination 5 (see Figure 2) which, despite consisting only of diagnoses, sex and age achieves a best precision of 0.933 and manages to recover a total of 605 new patients (25%) in OUH. In the case of Chelwest, the best precision for combination 5 is 1 and the percentage of recovery patients is of 3%.

Other feature combinations such as number 13 (Jaccard distance with Sex, P+S Diag, New medication names and P+S Procedures) also achieves a high best precision (0.806 for OUH and 1 for Chelwest) and recovers 561 new patients (23%) and 742 new patients (42%) for OUH and Chelwest respectively. Even though combination 13 performs better than combination 5 in Chelwest, we favour combination 5 since it achieves higher precisions with a smaller number of features while allowing for a significant increase in patient number.

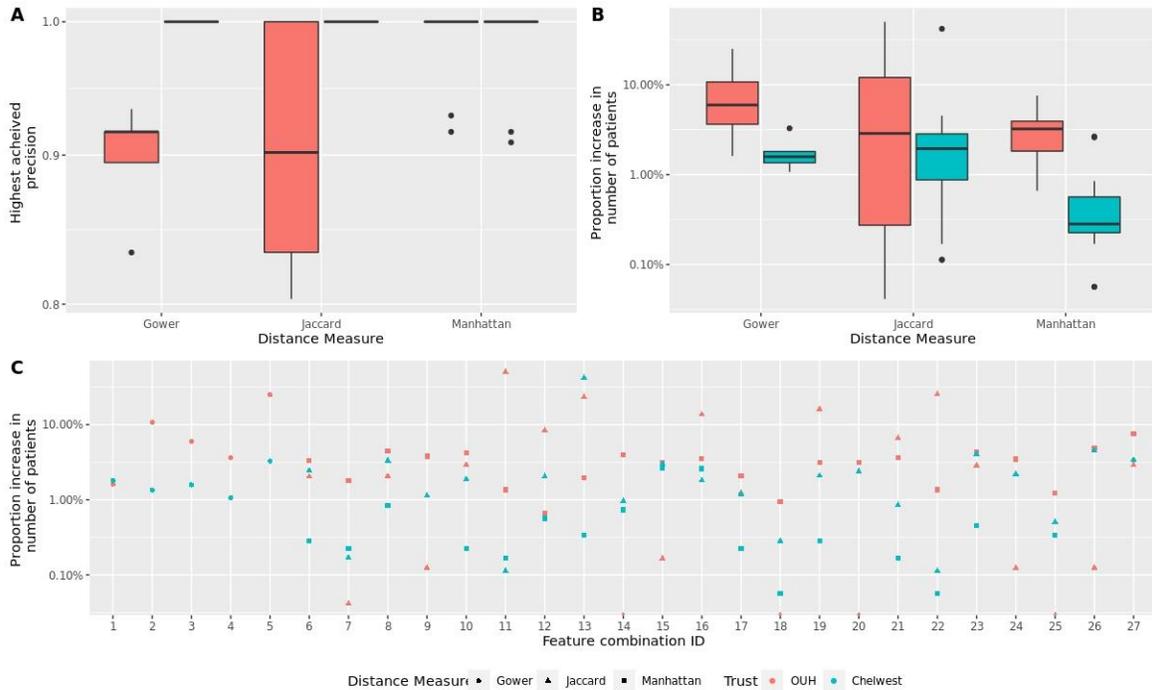

*Figure 3: Analysis of results from FILL algorithm applied to all feature combinations tested (see Figure 2). Panel A summarises the precisions achieved by FILL in all combinations considered with each distance measure (left-side of the distance corresponds to OUH and right side is Chelwest). Panel B summarises the number of new HFpEF patients identified by FILL, presented as percentage with respect to number of patients with available EF measurements. Panel C shows the results of in panel B split out by feature combination (Y axes in log-scale). Results have been obtained using the optimal hyperparameter combination that maximises the precision subject to a minimum of 10 true positives from a leave-one-out analysis.*

While focusing on the best achievable precision is important to understand the best achievable performances in real word scenarios, generally, it is sufficient to achieve a good accuracy while being able to obtain a significant number of new patients. Therefore, we performed an additional analysis to identify the parameters that maximise the number of true positives from a leave-one-out analysis, given that the precision is above a pre-set threshold (Figure 4); thereby, slightly sacrificing precision in return for more patients. Using this latter method with a threshold of 0.85 still results in a higher precision than classical machine learning approaches (Table 2).

Figure 4 shows the newly classified HFpEF patients identified by applying this approach to FILL. Results are shown only for feature combinations 1-5 which uses Gower distance, which previous analysis indicated provides very good performance. Feature combination 5 for Chelwest can extend the number of patients with available EF measurements by over 100%, effectively doubling the patient cohort but exclusively with patients with a high likelihood of being HFpEF only. The optimal distance in combination 5 for Chelwest is of 0.88 and 1.48E-02 the p-value threshold is of 1.48E-02. Besides the distance and p-value for OUH is 0.86 and 3.32E-04 respectively. Note that these results are different from those of Figure 3-C since those of Figure 4 impose a precision of at least 0.85 whereas in the latter there is no precision imposed.

Notably, for all the models tested, with both sets of optimal hyperparameters, none of the patients that were originally classified as HFmrEF were predicted to be HFpEF, supporting the correct working of FILL.

In general, tuning hyperparameters S and T to maximises the precision (while restricting the analysis to pairs that lead to at least 10 true positives in a leave-one-out analysis as shown in Figure 3) often results in small S and/or T, thus requiring smaller, more uniform neighbourhood for the imputation. However, this method may be too strict and result in insufficient numbers of new patients (Figure 3C and Figure 5). To further investigate this, we plotted the precision versus the number of patients discovered using combination 5 (Figure 5). The different dots in Figure 5 correspond to the best performance given a precision above 0.80, 0.85, 0.90, and 0.95. By doing this we allow FILL to find the best combination (in terms of p-values and distance) that maximizes the true positives therefore recovering the percentage of new patients for each threshold. As hypothesized, there is a negative relationship between the two which appears to be linear. For comparison, we also include the performance of the model without any precision threshold displayed as a diamond. Once again, feature combination 5 performs significantly well suggesting that this feature set contains sufficient information to identify new HFpEF patients.

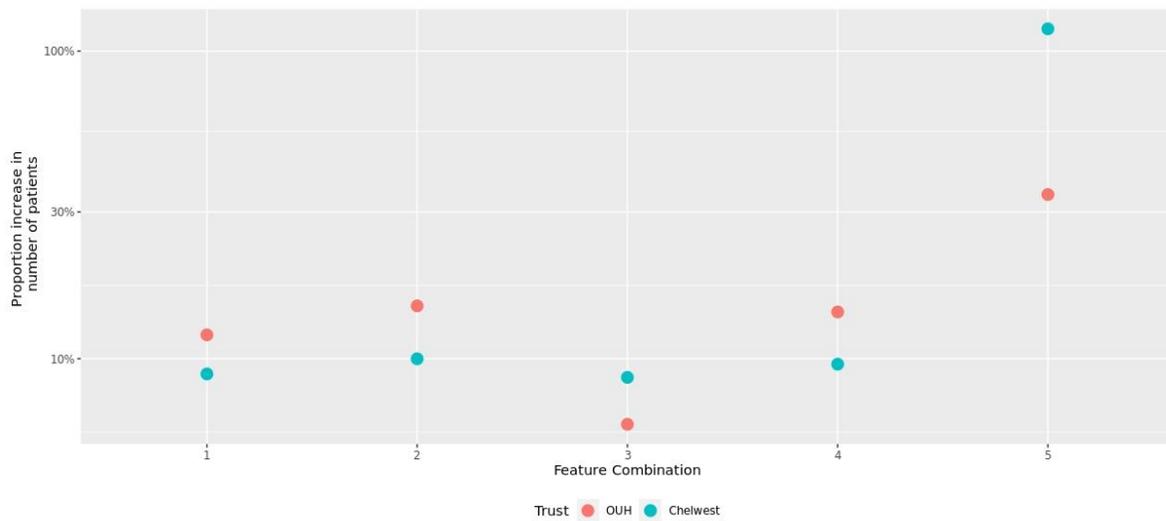

*Figure 4: Number of new HFpEF patients identified by FILL using the set of hyperparameters S and T that maximises the number of true positives from a leave-one-out analysis such that the precision is at least 0.85. Results are shown for only feature combinations (see Figure 2) that include age and therefore are all using Gower distance.*

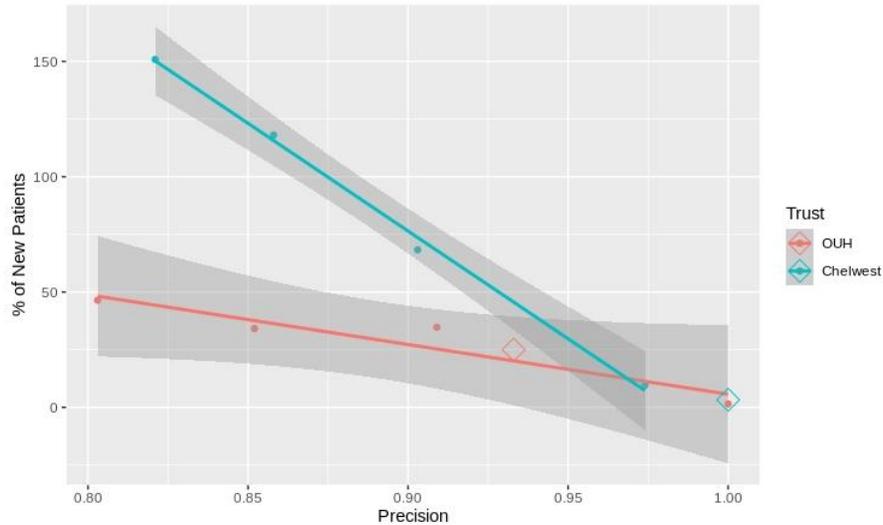

*Figure 5: Precision and percentage of new patients classified for combination 5 and Gower distance. Smooth linear trend has been introduced for each cohort connecting each point except the best performing one (without any cut off and indicated with an empty diamond).*

It has been suggested that HFpEF consists of several overlapping clinical phenotypes, which makes it challenging to classify a patient as HFpEF (Kao *et al.*, 2015; Uijl *et al.*, 2021). As FILL is a local-based method it is capable of dealing with potentially different subgroups of HFpEF patients. Furthermore, FILL allows for explainability: a newly HFpEF classified patients neighbours can be identified and the distribution of the features of the neighbours can be compared to the feature distribution of patients outside the neighbourhood. It is therefore possible, for each newly HFpEF classified patient, to distinguish a subset of features that makes their local neighbourhood have a higher-than-expected proportion of HFpEF patients.

To test the explainability of FILL, we explore the differences between a classified patient's neighbourhood and their respective non-neighbouring patients. We focus on results using data from OUH but similar findings were found for Chelwest which can be found in the supplementary material section. In order to do that, we employed a leave-one-out approach to classify patients with known EF using Gower distance on age, sex, and diagnoses. Using hyperparameters optimization given a precision of at least 0.85 as in Figure 4, FILL identified 46 true positives and 8 false positives with a precision of 0.852 for OUH (while 362 true positives and 60 false positives with a precision of 0.858 for Chelwest). We then randomly selected 8 true positives (HFpEF patients) and a false positive (HFrEF patient) and plotted the statistically significant odd ratios (for binary features) and difference in mean (for the age) when comparing neighbours VS non-neighbours patients (Figure 6A). At a first approximation, a larger number of statistically significant features (indicates in the panel title) indicates a neighbourhood with features that are very different from those of the rest of the patients.

Figure 6 indicates that some patients sit in rather unique neighbourhood, e.g., patient 7 with 160 features, suggesting the presence of cluster of patients with unique combination of features that are associated HFpEF patients. At the same time, some patient neighbourhoods only differ marginally to their non-neighbouring patients e.g., the neighbourhood around patient 6, suggesting more fuzzy clusters.

To further explore these results visually, we projected all the patients with known EF into a 2d manifold using UMAP (Figure 7). In this plot, each dot represents a patient (colour coded according to their EF status). The patients represented in Figure 6 are also indicated with their respective number. Although, a weak grouping seems to emerge in the bottom left corners, the randomly selected patients are rather spread out. Interestingly, patient 9 tends to be surrounded by the rest of patients, shedding some light on the reasons of the misclassification. All in all, this representation indicates that FILL is capable of identifying local areas of feature overrepresentations (i.e. HFpEF patients) there are not easily detectable using more visual approaches based on dimensionality reduction. Furthermore, the *fuzzy clusters* identified by FILL appear to be rather independent from clusters that would have been identified using a more classical analysis based on dimensionality reduction and unsupervised cluster detection.

In order to understand which features are most characteristic of the neighbourhoods considered, we explore the 5 most significant features (i.e., those with the smallest p-values) in Table 3 as well as a 2D UMAP space (Figure 7) for the 9 patients of Figure 6. Patients 5, 6, 2, and 9 which are grouped closer visually (left bottom corner) share as their first most significant feature the Z92.1 ICD10 code: "personal history of long-term use of anticoagulants". While in contrast patient 1 and 4, the patients which are furthest away in this space, don't have any of the most significant features in common.

Overall, and somehow unsurprisingly, ICD-10 subcodes of I48 (I48.X, "atrial fibrillation and flutter", and I48.9, "atrial fibrillation and flutter, unspecified") as well as I50 (I50.0, "heart failure", and I50.9, "hearth failure, unspecified") appear quite frequently across the 5 most significant features. More interestingly, several Z* ICD-10 codes ('Factors influencing health status and contact with health services') appear with a certain frequency, suggesting that the patient's history is a key factor in identifying HFpEF patients, of note is Z92.1 which is to be expected as treatment for HF patients.

Figure 6: Volcano plots of features for a sample of 9 patients from OUH that FILL classified as HFpEF when the feature set consisted of age, sex, and diagnoses. Results shown are from a leave-one-out analysis of patients with known EF status. Optimal hyperparameters used maximise the number of true positives given that the precision is >0.85. The feature set of HFpEF classified patients neighbouring patients were compared to non-neighbouring patients. Data is presented as odds ratio vs p-value for binary features and difference in mean value vs p-value for continuous features. FDR adjustment was used for p-values to adjust for multiple comparisons. Only features with an adjusted p-value below 0.05 are shown.

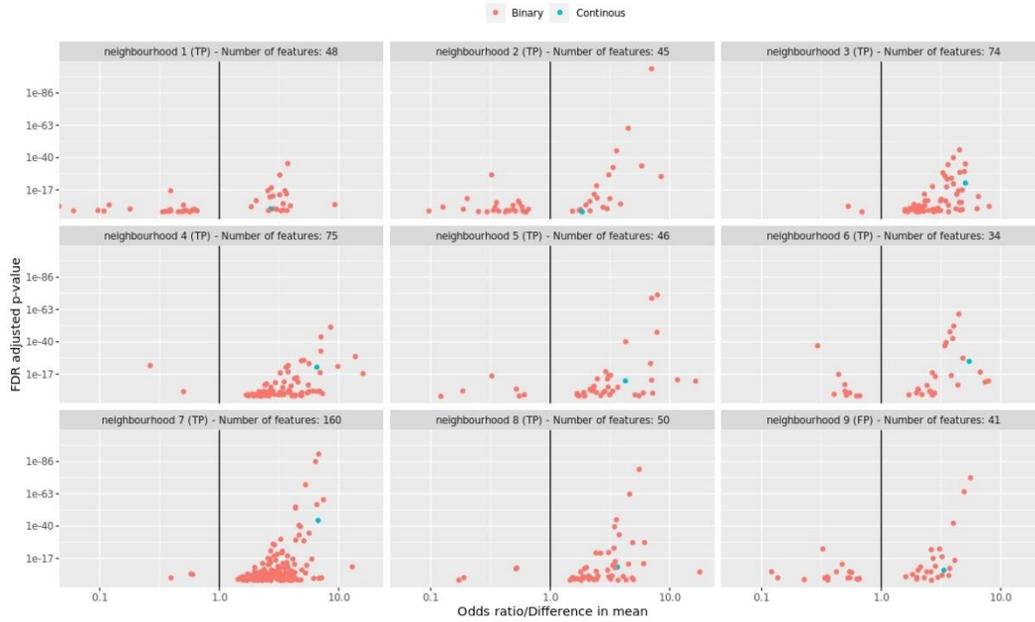

*Figure 6: Volcano plots of features for a sample of 9 patients from OUH that FILL classified as HFpEF when the feature set consisted of age, sex, and diagnoses. Results shown are from a leave-one-out analysis of patients with known EF status. Optimal hyperparameters used maximise the number of true positives given that the precision is >0.85. The feature set of HFpEF classified patients neighbouring patients were compared to non-neighbouring patients. Data is presented as odds ratio vs p-value for binary features and difference in mean value vs p-value for continuous features. FDR adjustment was used for p-values to adjust for multiple comparisons. Only features with an adjusted p-value below 0.05 are shown.*

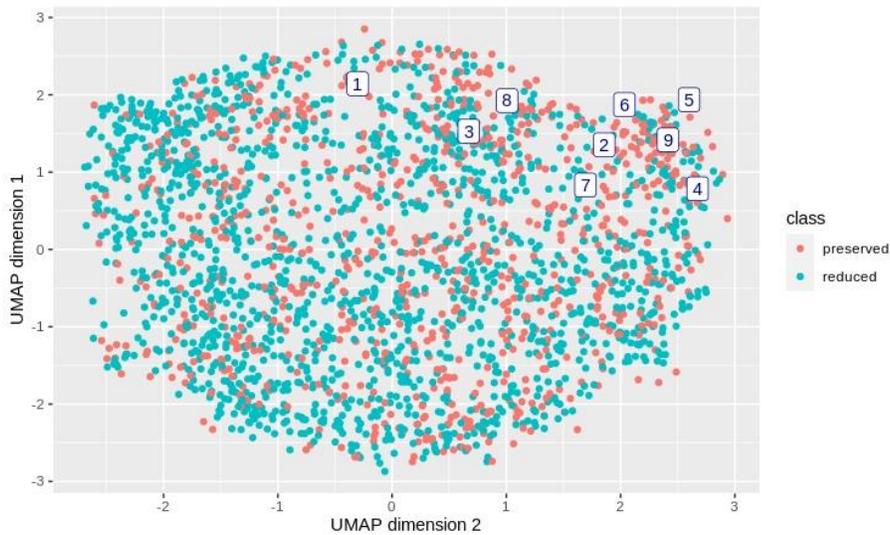

*Figure 7: UMAP of the know patients with combination 5 (Age + Sex + PS Diagnosis) for OUH. The dark-blue numbers correspond to the sample of 9 patients.*

Table 3: Summary of features found to be significantly different between neighbouring patients and non-neighbouring patients in a sample of 9 patients from OUH that FILL classified as HFpEF when the feature set consisted of age, sex, and diagnoses. Results shown are from a leave-one-out analysis of patients with known EF status. Optimal hyperparameters used maximise the number of true positives given that the precision is >0.85. The top 5 most significant features (as determined by FDR-adjusted p-value) for each patient neighbourhood ID are displayed with their corresponding odds ratio in brackets.

| ID | Most statistically significant features OUH | | | | |
|---|---|---|---|---|---|
| | 1st | 2nd | 3rd | 4th | 5th |
| 1 | I48.9 (OR 3.74) | Z92.1 (OR 3.21) | Z86.4 (OR 2.73) | isFemale (OR 2.55) / isMale (OR 0.392) | H26.9 (OR 3.55) |
| 2 | Z92.1 (OR 7.05) | I48.9 (OR 4.5) | I50.9 (OR 3.59) | Z95.4 (OR 5.82) | I48.X (OR 3.36) |
| 3 | Z86.4 (OR 4.52) | Z86.7 (OR 4.01) | Z86.6 (OR 5.06) | I48.9 (OR 3.62) | I10.X (OR 4.97) |
| 4 | Z50.7 (OR 8.56) | Z50.1 (OR 7.09) | M19.9 (OR 7.08) | M19.99 (OR 13.78) | N39.0 (OR 5.11) |
| 5 | Z92.1 (OR 7.9) | I48.9 (OR 7.07) | Z95.4 (OR 7.82) | I50.9 (OR 4.28) | Z95.2 (OR 6.91) |
| 6 | Z92.1 (OR 4.46) | I48.9 (OR 4.04) | I50.0 (OR 3.76) | I48.X (OR 3.97) | Z86.7 (OR 3.49) |
| 7 | I48.9 (OR 6.8) | Z92.1 (OR 6.4) | I50.0 (OR 5.28) | Z50.7 (OR 7.45) | R29.6 (OR 6.57) |
| 8 | Z92.1 (OR 5.56) | I48.9 (OR 4.62) | I50.9 (OR 3.58) | Z86.7 (OR 3.45) | I10.X (OR 3.78) |
| 9 | Z92.1 (OR 5.59) | I48.9 (OR 4.94) | I48.X (OR 4.01) | isMale (OR 3.08) / isFemale (OR 0.32) | I50.9 (OR 2.62) |

## 4. Discussion & Conclusion:

In this paper we have described a novel imputation approach names FILL. Unlike classical machine learning algorithms that aim to predict all missing data, FILL aims at imputing a value only for those patients that we can be associated with a specific value with a high likelihood. FILL is, therefore, able to confidently extend cohorts to supplement further analysis. In applying FILL to classify HFpEF patients when EF measurements are not known, we have shown that FILL is able to identify potential HFpEF patients with high accuracy and that simple approaches can be used to complement this analysis by providing information on the specific features used to support the algorithmic decision.

EPR are often rich in features, and it can be hard to decide which features need to be included when performing specific analyses. On the one hand, more features help better characterizing patients' medical condition, on the other hand using too many features may result in underpowered analyses which end up being less effective than expected. While approaches like autoencoders may ameliorate this problem (Carr *et al.*, 2021) there are situations in which this is challenging or even not possible.

Interestingly, our analysis suggests that a combination of demographics features and diagnoses proved to be quite powerful in classifying HFpEF patients (Figure 2-4). The addition of procedures, medications or laboratory measurements did not significantly improve the results. Overall, this is an encouraging result that indicate that limited amount of data may be enough to characterize, and hence potentially identify, HFpEF patients.

EPR data can be extremely noisy (Russell, 2021), therefore, clinically-guided feature curation is sometimes needed to help reduce this noise. Figure 3 demonstrates, for medication name, that clinically-guided feature curation was able to improve the results by removing genuine noise in the free-text medication name entered in the data. However, despite some noise within the data, FILL was able to perform extremely well in identifying patients with a high likelihood of being HFpEF (Figure 3-4). It is expected that further feature curation e.g., further cleaning of medication names, or aggregation of certain ICD-10 or OPCS-4 codes into disease groups may improve the results further. Another method of including clinical knowledge may be incorporated by giving features that are known to be associated with HFpEF more weight in the calculation of the distance measures.

In our current set up, we have treated the history of the patient in a static form, i.e., whether the patient has ever had a given diagnosis at any point in their EPR or not. While we have seen that this method is able to achieve relatively high precision levels, further refinement such as the use of time-window filtering for features around an index event of interest, e.g., HF diagnosis of a patient, may improve the results. An alternative approach would be to construct time-embeddings for the patient trajectory using an autoencoder (Carr *et al.*, 2021).

As previously discussed, a small number of HF patients (39 in OUH and 241 in Chelwest) with records of HFmrEF were considered as having unknown EF due to 1) concerns on their representativeness and 2) the absence of specific treatment guidelines for these patients in the NICE guidelines. None of these patients were predicted as being HFpEF in any of the models tested, supporting the correct working of FILL.

FILL works by identifying neighbouring patients who are similar to the patient being classified. Each of these neighbours then has equal influence on the classification of the patient, regardless of their level of similarity. One potential extension of the algorithm would be to weight the contributions of the neighbours by their distance from the patient being classified. It makes sense for neighbours who are closer (i.e., more similar) to the patient being classified to have more influence on the classification than neighbours who are more dissimilar and closer to the edge of the neighbourhood. One possible method for doing this would be to use the inverse distance as a weighting, giving neighbours who are closer (smaller distance and so more similar) more weight than neighbours who are further away (larger distance and so more dissimilar).

HFpEF is notoriously difficult to classify, both from a clinical (Huis in 't Veld *et al.*, 2016; Naing *et al.*, 2019) and an EHR prediction (Austin *et al.*, 2013; Ho *et al.*, 2016; Uijl *et al.*, 2020) standpoint. One potential explanation for this is that HFpEF is likely to consists of several different overlapping syndromes and, therefore, two HFpEF patients may present very differently in the clinic (Kao *et al.*, 2015; Uijl *et al.*, 2021). This diversity makes it challenging for global-based machine learning methods to identify, and characterize, these patients. The heterogeneity within HFpEF can be overcome by using more local-based methods, such as FILL, which are able to explore patients (dis-)similarity at a more local scale.

One limitation of the study is that the initial cohort was defined using ICD10 codes. We can expect human errors in labelling the diseases, which results in either additional patients who are falsely included in the study or patients with HF who are missed in the analysis. One way of addressing this problem is to include further laboratory measurements in the cohort definition such as BNP and EF to validate the diagnoses codes. However, these values are often only available for a limited number of patients (see motivation of this paper). Further, several clinicians indicated there is professional scepticism with the label of HFpEF, and most expressed a need for more knowledge and understanding of the importance of distinguishing this as an entity (Sowden *et al.*, 2020), resulting in noisy labels of our analysis. To address these limitations is outside the scope of this paper and remains an important research area.

Despite our analysis being restricted to only two NHS trusts, notable differences can be observed. Overall, the precision of FILL for Chelwest tends to be higher than for OUH, but given the inverse relationship between precision and number of new patients. The FILL algorithm tends to achieve higher proportion of new patients for OUH. This is illustrated in Figure 3C where for most of the 27 different combinations and three different distance measurements, there are more new patients classified in OUH than Chelwest. Nonetheless, for each unit of precision sacrificed, the proportion of new patients classified as HFpEF increases at a faster rate using Chelwest data compared to OUH data. There may be several reasons for these differences, one such reason could be the presence of as different coding standardisations in each Trust. However, FILL works well in both Trusts, supporting the idea that the algorithm can adapt to different situations.

Finally, it's worth pointing out that FILL allows for easy explainability as to why a patient has been classified as HFpEF by comparing the feature set of the local neighbours to those that are not within the neighbourhood. This allow for *accountability* in the results of the analysis, which is critical in clinical settings, where clinical decisions need to be supported by evidence and combined with clinical judgment in order to provide the best care for a patient.

5. Acknowledgments:

This work uses data provided by patients collected by Chelsea and Westminster Hospital NHS Foundation Trust and Oxford University Hospitals NHS Foundation Trust as part of their care and support. We believe using the patient data is vital to improve health and care for everyone and would, thus, like to thanks all those involved for their contribution. The data were extracted, anonymised, and supplied by the Trust in accordance with internal information governance review, NHS Trust information governance approval, and the General Data Protection Regulation (GDPR) procedures outlined under the Strategic Research Agreement (SRA) and relative Data Processing Agreements (DPAs) signed by the Trust and Sensyne Health plc.

This research has been conducted using the Oxford University Hospitals NHS Foundation Trust Clinical Data Warehouse, which is supported by the NIHR Oxford Biomedical Research Centre and Oxford University Hospitals NHS Foundation Trust. Special thanks to Kerrie Woods, Kinga V́arnai, Oliver Freeman, Hizni Salih, Steve Harris and Professor Jim Davies.

## 6. References:

Austin, P.C. *et al.* (2013) 'Using methods from the data-mining and machine-learning literature for disease classification and prediction: a case study examining classification of heart failure subtypes', *Journal of Clinical Epidemiology*, 66(4), pp. 398–407. doi:10.1016/j.jclinepi.2012.11.008.

Beaulieu-Jones, B.K. and Moore, J.H. (2017) 'MISSING DATA IMPUTATION IN THE ELECTRONIC HEALTH RECORD USING DEEPLY LEARNED AUTOENCODERS', *Pacific Symposium on Biocomputing. Pacific Symposium on Biocomputing*, 22, pp. 207–218. doi:10.1142/9789813207813_0021.

Blecker, S. *et al.* (2016) 'Comparison of Approaches for Heart Failure Case Identification From Electronic Health Record Data', *JAMA cardiology*, 1(9), pp. 1014–1020. doi:10.1001/jamacardio.2016.3236.

Bloom, B.M. *et al.* (2021) 'Usability of electronic health record systems in UK EDs', *Emergency Medicine Journal*, 38(6), pp. 410–415. doi:10.1136/emermed-2020-210401.

Borlaug, B.A. (2020) 'Evaluation and management of heart failure with preserved ejection fraction', *Nature Reviews Cardiology*, 17(9), pp. 559–573. doi:10.1038/s41569-020-0363-2.

Carr, O. *et al.* (2021) 'Longitudinal patient stratification of electronic health records with flexible adjustment for clinical outcomes', in *Proceedings of Machine Learning for Health. Machine Learning for Health*, PMLR, pp. 220–238. Available at: https://proceedings.mlr.press/v158/carr21a.html (Accessed: 17 December 2021).

Chan, Y.H. *et al.* (2003) 'Vaccine Clinical Trials', in *in Encyclopedia of Biopharmaceutical Statistics, 2nd Edition, Marcel Dekker*, pp. 1005–1022.

Gower, J.C. (1971) 'A General Coefficient of Similarity and Some of Its Properties', *Biometrics*, 27(4), pp. 857–871. doi:10.2307/2528823.

Heidenreich, P.A. *et al.* (2013) 'Forecasting the impact of heart failure in the United States: a policy statement from the American Heart Association', *Circulation. Heart Failure*, 6(3), pp. 606–619. doi:10.1161/HHF.0b013e318291329a.

Ho, J.E. *et al.* (2016) 'Predicting Heart Failure With Preserved and Reduced Ejection Fraction: The International Collaboration on Heart Failure Subtypes', *Circulation. Heart Failure*, 9(6), p. e003116. doi:10.1161/CIRCHEARTFAILURE.115.003116.

Huis in 't Veld, A.E. *et al.* (2016) 'How to diagnose heart failure with preserved ejection fraction: the value of invasive stress testing', *Netherlands Heart Journal*, 24(4), pp. 244–251. doi:10.1007/s12471-016-0811-0.

Jaccard, P. (1912) 'The Distribution of the Flora in the Alpine Zone.1', *New Phytologist*, 11(2), pp. 37–50. doi:10.1111/j.1469-8137.1912.tb05611.x.

Jafari, M. and Ansari-Pour, N. (2019) 'Why, When and How to Adjust Your P Values?', *Cell Journal (Yakhteh)*, 20(4), pp. 604–607. doi:10.22074/cellj.2019.5992.

Kao, D.P. *et al.* (2015) 'Characterization of subgroups of heart failure patients with preserved ejection fraction with possible implications for prognosis and treatment response', *European Journal of Heart Failure*, 17(9), pp. 925–935. doi:10.1002/ejhf.327.

Koye, D.N. *et al.* (2020) 'Temporal Trend in Young-Onset Type 2 Diabetes—Macrovascular and Mortality Risk: Study of U.K. Primary Care Electronic Medical Records', *Diabetes Care*, 43(9), pp. 2208–2216. doi:10.2337/dc20-0417.

Lee, D.S. *et al.* (2009) 'Relation of disease pathogenesis and risk factors to heart failure with preserved or reduced ejection fraction: insights from the framingham heart study of the national heart, lung, and blood institute', *Circulation*, 119(24), pp. 3070–3077. doi:10.1161/CIRCULATIONAHA.108.815944.

Li, J. *et al.* (2021) 'Imputation of missing values for electronic health record laboratory data', *npj Digital Medicine*, 4(1), pp. 1–14. doi:10.1038/s41746-021-00518-0.

Lloyd-Jones, D.M. *et al.* (2002) 'Lifetime risk for developing congestive heart failure: the Framingham Heart Study', *Circulation*, 106(24), pp. 3068–3072. doi:10.1161/01.cir.0000039105.49749.6f.

McInnes, L., Healy, J. and Melville, J. (2020) 'UMAP: Uniform Manifold Approximation and Projection for Dimension Reduction', *arXiv:1802.03426 [cs, stat]* [Preprint]. Available at: http://arxiv.org/abs/1802.03426 (Accessed: 30 November 2021).

Naing, P. *et al.* (2019) 'Heart failure with preserved ejection fraction: A growing global epidemic', *Australian Journal of General Practice*, 48(7), pp. 465–471. doi:10.31128/AJGP-03-19-4873.

Pencina, M.J. and D'Agostino, R.B., Sr (2015) 'Evaluating Discrimination of Risk Prediction Models: The C Statistic', *JAMA*, 314(10), pp. 1063–1064. doi:10.1001/jama.2015.11082.

Russell, L.B. (2021) 'Electronic Health Records: The Signal and the Noise', *Medical Decision Making*, 41(2), pp. 103–106. doi:10.1177/0272989X20985764.

Simmonds, S.J. *et al.* (2020) 'Cellular and Molecular Differences between HFpEF and HFrEF: A Step Ahead in an Improved Pathological Understanding', *Cells*, 9(1), p. E242. doi:10.3390/cells9010242.

Sowden, E. *et al.* (2020) 'Understanding the management of heart failure with preserved ejection fraction: a qualitative multiperspective study', *British Journal of General Practice*, 70(701), pp. e880–e889. doi:10.3399/bjgp20X713477.

Tison, G.H. *et al.* (2018) 'Identifying heart failure using EMR-based algorithms', *International Journal of Medical Informatics*, 120, pp. 1–7. doi:10.1016/j.ijmedinf.2018.09.016.

Uijl, A. *et al.* (2020) 'A registry-based algorithm to predict ejection fraction in patients with heart failure', *ESC heart failure*, 7(5), pp. 2388–2397. doi:10.1002/ehf2.12779.

Uijl, A. *et al.* (2021) 'Identification of distinct phenotypic clusters in heart failure with preserved ejection fraction', *European Journal of Heart Failure*, 23(6), pp. 973–982. doi:10.1002/ejhf.2169.

Wells, B.J. *et al.* (2013) 'Strategies for handling missing data in electronic health record derived data', *EGEMS (Washington, DC)*, 1(3), p. 1035. doi:10.13063/2327-9214.1035.

Wilcox, J.E. *et al.* (2020) 'Heart Failure With Recovered Left Ventricular Ejection Fraction: JACC Scientific Expert Panel', *Journal of the American College of Cardiology*, 76(6), pp. 719–734. doi:10.1016/j.jacc.2020.05.075.